\pdfoutput=1
\documentclass[conference]{IEEEtran}

\usepackage{
	amsmath,
	amssymb,
	booktabs,
	footnote,
	graphicx,
	multirow,
	pgfplots,
	subcaption,
	url,
    hyperref,
    siunitx,
    nicefrac,
    pifont,
    threeparttable,
    schemabloc,
    soul,
    verbatim,
    todonotes,
    float,
    textcomp
}
\usepackage[shortcuts]{extdash}

\usepackage{tikz}
\usetikzlibrary{arrows,shapes,circuits.logic.US,shapes.gates.logic.IEC,calc,decorations.markings,patterns}
\tikzstyle{branch}=[fill,shape=circle,minimum size=3pt,inner sep=0pt]

\usepgfplotslibrary{groupplots}

\usepackage[keeplastbox]{flushend}
\usepackage[graphicx]{realboxes}
\usepackage{stackengine}[2013-09-11]

\begin{document}

	\title{LUTNet: Rethinking Inference in FPGA Soft Logic}

	\author{
		\IEEEauthorblockN{Erwei Wang, James J. Davis, Peter Y. K. Cheung and George A. Constantinides}
		\IEEEauthorblockA{
			Department of Electrical and Electronic Engineering\\
			Imperial College London, London, SW7 2AZ, United Kingdom\\
			\texttt{\{erwei.wang13, james.davis, p.cheung, g.constantinides\}@imperial.ac.uk}
		}
	}

	\maketitle

	\begin{abstract}
	
		Research has shown that deep neural networks contain significant redundancy, and that high classification accuracies can be achieved even when weights and activations are quantised down to binary values.
		Network binarisation on FPGAs greatly increases area efficiency by replacing resource-hungry multipliers with lightweight XNOR gates.
		However, an FPGA's fundamental building block, the $K$-LUT, is capable of implementing far more than an XNOR: it can perform any $K$-input Boolean operation.
		Inspired by this observation, we propose LUTNet, an end-to-end hardware-software framework for the construction of area-efficient FPGA-based neural network accelerators using the native LUTs as inference operators.
		We demonstrate that the exploitation of LUT flexibility allows for far heavier pruning than possible in prior works, resulting in significant area savings while achieving comparable accuracy.
		Against the state-of-the-art binarised neural network implementation, we achieve twice the area efficiency for several standard network models when inferencing popular datasets.
		We also demonstrate that even greater energy efficiency improvements are obtainable.

	\end{abstract}

	\section{Introduction and Motivation}
	\label{sec:intro}
	
		During inference, the most common---and expensive---computational node in a deep neural network (DNN) performs a function of the form in \eqref{eq:synapse_normal}, calculating a channel output $y$.
		Each weight $w_n$ is a constant determined during training, $\boldsymbol{x}$ a vector of $N$ channel inputs and $f$ an activation function such as the widely used rectified linear unit.
		In the extreme case where $\boldsymbol{w} \in \left\{-1,1\right\}^N$---so-called binarised neural networks (BNNs)---the multiplications become cheap or free to implement.
		When time-multiplexed, multipliers become XNOR gates.
		When unrolled, they can be further simplified into buffers and inverters, all of which are usually subsumed into the downstream adder logic.
		Also beneficial for BNNs is the ability to use a population count (popcount) for the summation: an operation that consumes half the LUTs of the otherwise-throughput-optimal balanced adder tree~\cite{BNN_CNN_FINN}.
		
		\begin{equation}
		    y = f{\left(\sum_{n=1}^N{w_n x_n}\right)}
			\label{eq:synapse_normal}
		\end{equation}
		
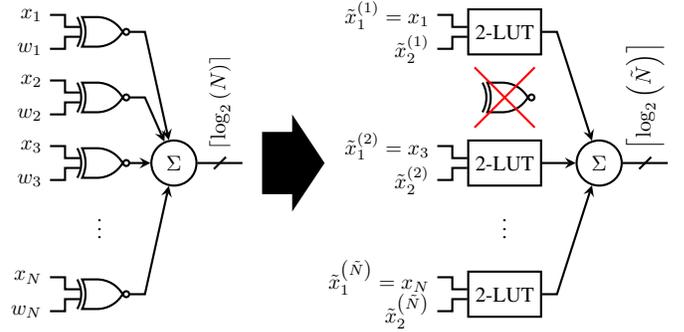
\begin{figure}
		    \centering
		    \begin{tikzpicture}[thick, circuit logic US, thick, node distance=2.5mm, label distance=2mm, decoration={markings, mark=at position 0.5 with {\node [font=\footnotesize] {/};}}, scale=0.8, every node/.style={scale=0.8}]
	
	\newcommand{\updownkinklen}{2mm}
	\newcommand{\leftkinklen}{2.5mm}
	\newcommand{\upperinputlen}{2.5mm}
	\newcommand{\lowerinputlen}{2.5mm}
	\newcommand{\rightkinklen}{2.5mm}
	\newcommand{\outputlen}{7.5mm}
	
	\tikzstyle {xnor} = [xnor gate, point right, inputs={nn}]
	\tikzstyle {lut} = [rectangle, minimum width=12mm, minimum height=7.5mm, align=center, draw]
	\tikzstyle {sum} = [circle, minimum width=7.5mm, minimum height=7.5mm, align=center, draw]
	\tikzstyle {arrow} = [->, >=stealth]
	
	\node (lut3) [lut] {2-LUT};
	\node (lut2) [lut, above=of lut3, white] {2-LUT};
	\node (pruned) [xnor] at (lut2) {};
	\node (lut1) [lut, above=of lut2] {2-LUT};
	\node (lut4) [lut, below=of lut3, white] {};
	\node at ([yshift=1mm]lut4) {$\vdots$};
	\node (lut5) [lut, below=of lut4] {2-LUT};
	\node (sum2) [sum, right=of lut3, xshift=2.5mm] {$\Sigma$};
	
	\node (to) [left=of lut3, xshift=-24.5mm] {};
	\draw [fill=black] ([xshift=-5mm, yshift=-5mm]to.center) -- ([xshift=-5mm, yshift=5mm]to.center) -- ([yshift=5mm]to.center) -- ([yshift=7.5mm]to.center) -- ([xshift=5mm]to.center) -- ([yshift=-7.5mm]to.center) -- ([yshift=-5mm]to.center) -- cycle;
	
	\node (sum1) [sum, left=of to, xshift=-11.5mm] {$\Sigma$};
	\node (xnor3) [xnor, left=of sum1, xshift=-2.5mm] {};
	\node (xnor1) [xnor] at (lut1 -| xnor3) {};
	\node (xnor2) [xnor] at (lut2 -| xnor3) {};
	\node at ([yshift=1mm]lut4 -| xnor3) {$\vdots$};
	\node (xnor5) [xnor] at (lut5 -| xnor3) {};
	
	\draw [red] ([xshift=-5mm, yshift=-5mm]pruned.center) -- ([xshift=5mm, yshift=5mm]pruned.center);
	\draw [red] ([xshift=-5mm, yshift=5mm]pruned.center) -- ([xshift=5mm, yshift=-5mm]pruned.center);
	
	\draw (xnor1.input 1 -| lut1.west) -- ++(left:\leftkinklen) -- ++(up:\updownkinklen) -- ++(left:\upperinputlen) node [left, anchor=east] {$\tilde{x}_1^{\left(1\right)} = x_1$};
	\draw (xnor1.input 2 -| lut1.west) -- ++(left:\leftkinklen) -- ++(down:\updownkinklen) -- ++(left:\lowerinputlen) node [left, anchor=east] {$\tilde{x}_2^{\left(1\right)}$};
	\draw (xnor3.input 1 -| lut3.west) -- ++(left:\leftkinklen) -- ++(up:\updownkinklen) -- ++(left:\upperinputlen) node [left, anchor=east] {$\tilde{x}_1^{\left(2\right)} = x_3$};
	\draw (xnor3.input 2 -| lut3.west) -- ++(left:\leftkinklen) -- ++(down:\updownkinklen) -- ++(left:\lowerinputlen) node [left, anchor=east] {$\tilde{x}_2^{\left(2\right)}$};
	\draw (xnor5.input 1 -| lut5.west) -- ++(left:\leftkinklen) -- ++(up:\updownkinklen) -- ++(left:\upperinputlen) node [left, anchor=east] {$\tilde{x}_1^{\left(\tilde{N}\right)} = x_N$};
	\draw (xnor5.input 2 -| lut5.west) -- ++(left:\leftkinklen) -- ++(down:\updownkinklen) -- ++(left:\lowerinputlen) node [left, anchor=east] {$\tilde{x}_2^{\left(\tilde{N}\right)}$};
	\draw [arrow] (lut1.east) -- ++(right:\rightkinklen) -- (sum2);
	\draw [arrow] (lut3.east) -- ++(right:\rightkinklen) -- (sum2);
	\draw [arrow] (lut5.east) -- ++(right:\rightkinklen) -- (sum2);
	\draw (sum2.east) -- ++(right:\outputlen);
	\draw ([xshift=-1mm, yshift=-1mm]$(sum2.east) + (\outputlen/2,0)$) -- ([xshift=1mm, yshift=1mm]$(sum2.east) + (\outputlen/2,0)$) node [midway, above, anchor=west, rotate=90, xshift=1mm] {$\left\lceil\log_2{\left(\tilde{N}\right)}\right\rceil$};
	
	\draw (xnor1.input 1) -- ++(left:\leftkinklen) -- ++(up:\updownkinklen) -- ++(left:\upperinputlen) node [left, anchor=east] {$x_1$};
	\draw (xnor1.input 2) -- ++(left:\leftkinklen) -- ++(down:\updownkinklen) -- ++(left:\lowerinputlen) node [left, anchor=east] {$w_1$};
	\draw (xnor2.input 1) -- ++(left:\leftkinklen) -- ++(up:\updownkinklen) -- ++(left:\upperinputlen) node [left, anchor=east] {$x_2$};
	\draw (xnor2.input 2) -- ++(left:\leftkinklen) -- ++(down:\updownkinklen) -- ++(left:\lowerinputlen) node [left, anchor=east] {$w_2$};
	\draw (xnor3.input 1) -- ++(left:\leftkinklen) -- ++(up:\updownkinklen) -- ++(left:\upperinputlen) node [left, anchor=east] {$x_3$};
	\draw (xnor3.input 2) -- ++(left:\leftkinklen) -- ++(down:\updownkinklen) -- ++(left:\lowerinputlen) node [left, anchor=east] {$w_3$};
	\draw (xnor5.input 1) -- ++(left:\leftkinklen) -- ++(up:\updownkinklen) -- ++(left:\upperinputlen) node [left, anchor=east] {$x_N$};
	\draw (xnor5.input 2) -- ++(left:\leftkinklen) -- ++(down:\updownkinklen) -- ++(left:\lowerinputlen) node [left, anchor=east] {$w_N$};
	\draw [arrow] (xnor1.output) -- ++(right:\rightkinklen) -- (sum1);
	\draw [arrow] (xnor2.output) -- ++(right:\rightkinklen) -- (sum1);
	\draw [arrow] (xnor3.output) -- ++(right:\rightkinklen) -- (sum1);
	\draw [arrow] (xnor5.output) -- ++(right:\rightkinklen) -- (sum1);
	\draw (sum1.east) -- ++(right:\outputlen);
	\draw ([xshift=-1mm, yshift=-1mm]$(sum1.east) + (\outputlen/2,0)$) -- ([xshift=1mm, yshift=1mm]$(sum1.east) + (\outputlen/2,0)$) node [midway, above, anchor=west, rotate=90, xshift=1mm] {$\left\lceil\log_2{\left(N\right)}\right\rceil$};
	
\end{tikzpicture}
		    \caption{
		        BNN to LUTNet architectural transformation for a single channel, mirroring the replacement of \eqref{eq:synapse_normal} with \eqref{eq:synapse_lutnet}.
		        Activation function blocks are not shown, but follow the adders.
		        $\tilde{N}$ lookup tables (here, 2-LUTs) substitute $N$ XNOR gates. $\tilde{N} \ll N$ is achieved via pre-substitution pruning, represented by the removal---\emph{i.e.} lack of LUT substitution---of the second XNOR gate.
		        LUT inputs $\tilde{x}_1^{\left(n\right)}~\forall n$ are connected to preserve the pruned BNN's structure.
		        LUTNet's weights are encoded in its LUT masks, thus do not appear as inputs.
            }
		    \label{fig:transform}
		\end{figure}

		No matter how simple these multiplications become, however, all of the products still need to be summed.
		In modern networks, $N$ commonly reaches numbers in the thousands~\cite{MOBILENET, ALEXNET}.
		To tackle this, we propose the replacement of \eqref{eq:synapse_normal} with the specifically FPGA-inspired function \eqref{eq:synapse_lutnet}, wherein the activation function is unchanged but each product is replaced with an \emph{arbitrary} term-specific Boolean function $g_n : \left\{-1,1\right\}^K \to \left\{-1,1\right\}$.
		The input to this function is a vector $\tilde{\boldsymbol{x}}^{\left(n\right)}$ whose elements are any $K$ components of the original input vector $\boldsymbol{x}$, \emph{i.e.} $\tilde{\boldsymbol{x}}^{\left(n\right)} = \boldsymbol{S}_n\boldsymbol{x}$ for some binary selection matrix $\boldsymbol{S}_n \in \left\{0,1\right\}^{K \times N}$ with ${\left\lVert\boldsymbol{S}_n\right\rVert}_\infty = 1$.
		Since its inputs and outputs are binary, each $g_n$ maps directly to a single $K$-LUT.
		BNNs are a special case of this function: they are recoverable for $K = 1$ and $\tilde{N} = N$, with $\boldsymbol{S}_n$ being the row vector with the $n$\textsuperscript{th} element equal to one and all others zero.
		An example of the resultant architectural transformation---excluding blocks for $f$, which are common to both approaches---is given in Fig.~\ref{fig:transform}.
		
		\begin{equation}
			y = f{\left(\sum_{n=1}^{\tilde{N}} g_n{\left(\tilde{\boldsymbol{x}}^{\left(n\right)}\right)}\right)}
			\label{eq:synapse_lutnet}
		\end{equation}
		
		Notice that, while in \eqref{eq:synapse_normal} each element of $\boldsymbol{x}$ only participates in a single summation term, in \eqref{eq:synapse_lutnet} each can participate in many terms.
		The intuition here is that inputs can be arranged such that $\tilde{N} \ll N$ for comparable accuracy via network pruning, dramatically reducing the sizes of the required popcount trees.
		Our experiments demonstrate that this is indeed the case.
		
		Our aim in proposing this inference node function is to play to the strengths of FPGA soft logic.
		While a LUT is capable of performing an arbitrary \emph{nonlinear Boolean} function, traditional DNNs are based around \emph{near-linear high-precision} functions: almost the exact opposite of the architecture's forte.
		Innovations such as BNNs have addressed one side of this weakness, by reducing precision~\cite{CSUR}; we address both by also embracing the nonlinearity of the LUT.
		
		In this paper, we make the following novel contributions:
		\begin{itemize}
			\item
			    We introduce LUTNet, the first neural network architecture featuring $K$-LUTs as inference operators.
			    Since each $K$-LUT is capable of performing an arbitrary Boolean operation on up to $K$ binary inputs, LUTNet's logic density is much greater than that of BNNs.
			\item
			    We propose a training regime resulting in the conversion of a BNN architecture from a dense array of simple XNOR gates into a sparse network of arbitrary $K$-input functions directly mappable onto $K$-LUTs.
			\item
			    We empirically demonstrate the effects of LUTNet's increased logic density on area efficiency and accuracy.
			    We also experimentally explore the associated energy and training efficiency impacts.
			    Our results for 4-LUT-based inference operators reveal area compression of 2.08$\times$ and 1.90$\times$ for the CNV network classifying the CIFAR-10 dataset and AlexNet classifying ImageNet, respectively, against an unrolled and losslessly pruned implementation of ReBNet~\cite{BNN_CNN_REBNET_FCCM}, the state-of-the-art BNN, with accuracy bounded within $\pm$0.300~percentage points (pp).
		\end{itemize}
		
	\section{Related Work}
	\label{sec:background}
		    
	    The authors of early BNN publications, such as BinaryConnect~\cite{BNN_CNN_BinaryConnect} and BinaryNet~\cite{BNN_CNN_BinaryNet}, proposed network training with binary weights and activations (channel inputs and outputs) used for forward propagation.
	    High-precision formats---most commonly IEEE-754 single-precision floating point, used to approximate  reals $\mathbb{R}$---are always used for backward propagation; this is essential in order for stochastic gradient descent to work well~\cite{BNN_CNN_BinaryConnect}.
        Tang \emph{et al.} showed that training from scratch with binarised forward propagation is significantly slower than through the consistent use of high-precision data, however; learning rates some 100$\times$ lower are required than in the all-real case~\cite{BNN_CNN_BINARY_CONSTRIANED_TRAINING}.
        Furthermore, binary forward propagation results in the majority of real-valued weights being close to either $-1$ or $1$, while a spread across $\left[-1, 1\right]$ is required to facilitate fine-grained pruning~\cite{PRU_CNN_TRAIN_PRUNE_RETRAIN}.
        
    	Use of fine-grained pruning effectively adds zero to the set of possible weight values, resulting in a ternary representation.
    	Ternarisation has been shown by the authors of many works to deliver significantly higher accuracy than yielded through binarisation~\cite{TNN_CNN_TWN,TNN_CNN_TTQ,TNN_CNN_BENGIO}.
        Pruning also promotes regularisation, reducing overfitting~\cite{PRU_CNN_STRUCTURED_SPARSITY}.
        The latter is particularly relevant to this work since the use of $K$-LUTs as inference operators greatly increases potential network complexity.
        
        In order to promote pruning, Han \emph{et al.} proposed training with the $l_2$ sparsification regulariser in \eqref{eqn:cnn_sparsity_regulariser}~\cite{PRU_CNN_TRAIN_PRUNE_RETRAIN}.
        During backward propagation, the value of $\Omega$ influences training loss, inducing weights carrying low significance to descend towards zero.
        $\lambda$, $L$ and $C$ are the regularisation factor, number of layers and number of channels per layer, respectively.
        $\hat{\boldsymbol{w}}^{\left(l,c\right)}$ denotes the real-valued weight vector of layer $l$'s channel $c$.
        
        \begin{equation}
            \Omega = \lambda\sqrt{\sum^L_{l=1}{\sum^C_{c=1}{{\left(\hat{\boldsymbol{w}}^{\left(l,c\right)}\right)}^2}}}
            \label{eqn:cnn_sparsity_regulariser}
        \end{equation}
        
        Improving upon BinaryNet's data representation, Rastegari \emph{et al.}'s BWN features layer-wise trainable scaling factors $\boldsymbol{\alpha}$ used in order to increase BNN expressiveness~\cite{BNN_CNN_XNOR-Net}.
        During training, each $\alpha_l \in \mathbb{R}$ assumes the mean value of layer $l$'s weights.
        When inferencing, this is multiplied with the layer's popcount results, compensating for some of the information lost to binarisation and increasing accuracy.

        Tang \emph{et al.}~\cite{BNN_CNN_BINARY_CONSTRIANED_TRAINING} and the authors of ABC-Net~\cite{BNN_CNN_ABC-Net} and ReBNet~\cite{BNN_CNN_REBNET_FCCM} demonstrated the alleviation of information loss from binarisation through the approximation of real-valued weights as linear combinations of multiple binary values.
        This is achieved via \emph{residual binarisation}, a scheme in which each bit is the binarised residual error of its predecessor.
        Each bit $b$ is associated with a trainable scaling factor $\gamma_b \in \mathbb{R}$, representing its relative importance.
        When quantising, each weight $\hat{w} \in \mathbb{R}$ is approximated as $B$ binary weights $w_b = \text{sign}{\left(\epsilon_b\right)}$, as shown in \eqref{eqn:residual_binarisation}, wherein $\epsilon_b$ is the $b$\textsuperscript{th} bit's residual error.
        During training, each $\gamma_b$ is updated to minimise the total error.
        While accuracy was found to be positively correlated with $B$, diminishing returns were seen; little improvement was observed for $B > 2$.
        
        \begin{equation}
            \begin{gathered}
                \hat{w} = \sum^B_{b=1}{\gamma_b~w_b} \\
                \epsilon_b = \epsilon_{b-1}-\gamma_{b-1}~\text{sign}{\left(\epsilon_{b-1}\right)}
            \end{gathered}
            \label{eqn:residual_binarisation}
        \end{equation}
        
        The aforementioned proposals are complementary to our approach, thus we embrace all of them.
        Through the use of high-precision training, fine-grained pruning, layer-wise scaling factors and residual binarisation, LUTNet achieves state-of-the-art accuracy.
        None of these lies at the heart of our proposal, however, and we do not consider their combination to represent novelty.
        Our novel use of $K$-LUTs allows us to reach such levels of performance significantly more cheaply than previously reported in the literature.
		
	\section{Network Construction and Training}
	\label{sec:training}
	    
	    LUTNet's initialisation comprises three successive stages: training, pruning and \emph{``logic expansion"} (XNOR to $K$-LUT conversion), with each of the latter two including a retraining phase.
	    These are shown enclosed within a dashed box in Fig.~\ref{plot:OVERALL_FLOW}.
	    All three phases were implemented with TensorFlow.
	    While our training and pruning stages are fairly standard, logic expansion encompasses the key novelty of our approach.
	    
	    \subsection{Training}
	    \label{sec:training_training}
            
            In order to both expedite learning and facilitate later pruning, our first step is to train the chosen network model using high-precision data during both forward and backward propagation.
            Layer-wise scaling factors $\alpha$ are learnt during this stage along with weights, and sparsification is induced through the use of the $l_2$ regulariser in \eqref{eqn:cnn_sparsity_regulariser} with $\lambda = \num{5e-7}$ as suggested by Tang \emph{et al.}~\cite{BNN_CNN_BINARY_CONSTRIANED_TRAINING}.
	    
	    \subsection{Pruning}
	    \label{sec:training_pruning}
	    
	        Following high-precision training, fine-grained pruning is conducted through the application of threshold $\theta$ on each weight $\hat{w}$, as shown in \eqref{eqn:threshold}.
	        The higher the value of $\theta$, the more weights are pruned away, exposing a continuum between area occupancy and accuracy.
	        
            \begin{equation}
                \begin{gathered}
                    \hat{w} \gets \begin{cases}
                        \hat{w} & \text{if}~\lvert \hat{w} \rvert > \theta \\
                        0       & \text{otherwise}
                    \end{cases}
                \end{gathered}
                \label{eqn:threshold}
            \end{equation}
	        
	        Once pruned, the network is binarised following the scheme shown in \eqref{eqn:residual_binarisation}, after which it is retrained in order to recover some of the induced accuracy loss.
	        Due to the diminishing returns previously found when applying residual binarisation~\cite{BNN_CNN_BINARY_CONSTRIANED_TRAINING,BNN_CNN_ABC-Net,BNN_CNN_REBNET_FCCM}, we used $B = 2$ (two-level binarisation) consistently.

	    \subsection{Logic Expansion}
	    \label{sec:training_expansion}
	    
	        At this point, we have obtained a residual-binarised ternary neural network with non-zero-weighted operators implemented as XNORs.
	        It is from here that we depart from the standard BNN approach.
	        Each XNOR gate is replaced with a $K$-LUT, whose first input $\tilde{x}^{\left(n\right)}_1$ is assigned to preserve the original connection, thereby retaining the pruned BNN's structure.
	        The $K - 1$ subsequent inputs to each LUT are then randomly selected from channel inputs within the same convolutional window as $\tilde{x}^{\left(n\right)}_1$, ensuring that the window shape remains unchanged.
	        We additionally constrain their selection such that each channel input is not multiply connected to the same LUT.
	        
	        The form of the inference function proposed in \eqref{eq:synapse_lutnet} is defined on the binary domain $\left\{-1,1\right\}^{\tilde{N}}$. 
            In common with quantisation-inspired networks, such as BNNs, this causes difficulty for training algorithms designed to operate on real vectors ${\mathbb R}^{\tilde{N}}$, specifically in the backward propagation of derivatives. 
            Our approach to this problem is to define an \emph{interpolating extension} of the function $g_n : \left\{-1,1\right\}^K \to \left\{-1,1\right\}$, \emph{i.e.} a function $\hat{g}_n : \mathbb{R}^K \to \mathbb{R}$ such that $\hat{g}_n{\left(\tilde{\boldsymbol{x}}^{\left(n\right)}\right)} = g_n{\left(\tilde{\boldsymbol{x}}^{\left(n\right)}\right)}$ for every $\tilde{\boldsymbol{x}}^{\left(n\right)}$ in the domain of $g_n$.
            There are many such functions.
            Of them, we prefer those that are as smooth as possible, allowing training optimisation methods to perform well, and that form a good interpolation in the sense that, if $g_n$ remains constant when a Boolean input flips, so does $\hat{g}_n$.
            A natural choice for the extension is a Lagrange interpolating polynomial, leading to the form we use in \eqref{eqn:Lagrange_interpl}.
            
            \begin{equation}
                \hat{g}_n{\left(\tilde{\boldsymbol{x}}^{\left(n\right)}\right)} = \sum_{\boldsymbol{d} \in \left\{-1,1\right\}^K}{\left(\hat{c}_{\boldsymbol{d}} \prod_{k = 1}^K{\left(\tilde{x}^{\left(n\right)}_k - d_k\right)}\right)}
                \label{eqn:Lagrange_interpl}
            \end{equation}
            
            \noindent This expands as shown in \eqref{eqn:Lagrange_interpl_expand} for $K \in \mathbb{N}_{>0}$, with each polynomial comprising $2^K$ trainable parameters $\hat{\boldsymbol{c}}$.
            
            \begin{equation}
                \hat{g}_n{\left(\tilde{\boldsymbol{x}}^{\left(n\right)}\right)} =
                \begin{cases}
                    \hat{c}_{\left(-1\right)}{\left(\tilde{x}^{\left(n\right)}_1 + 1\right)} + \hat{c}_{\left(1\right)}{\left(\tilde{x}^{\left(n\right)}_1 - 1\right)}	& \text{if}~K = 1	\\
                    \\
                    \begin{aligned}[c]
                        &\hat{c}_{\left(-1,-1\right)}{\left(\tilde{x}^{\left(n\right)}_1 + 1\right)\left(\tilde{x}^{\left(n\right)}_2 + 1\right)} \\
                        &+ \hat{c}_{\left(-1,1\right)}{\left(\tilde{x}^{\left(n\right)}_1 + 1\right)\left(\tilde{x}^{\left(n\right)}_2 - 1\right)} \\
                        &+ \hat{c}_{\left(1,-1\right)}{\left(\tilde{x}^{\left(n\right)}_1 - 1\right)\left(\tilde{x}^{\left(n\right)}_2 + 1\right)} \\
                        &+ \hat{c}_{\left(1,1\right)}{\left(\tilde{x}^{\left(n\right)}_1 - 1\right)\left(\tilde{x}^{\left(n\right)}_2 - 1\right)}
                    \end{aligned}	& \text{if}~K = 2   \\
                    \\
                    \cdots  & \cdots
                \end{cases}
                \label{eqn:Lagrange_interpl_expand}
			\end{equation}

	        Since connections are effectively remade from an unpruned BNN (Section~\ref{sec:training_training}), it makes sense to use those channel inputs' original weights as a starting point for retraining.
	        For each LUT, this is done by solving \eqref{eqn:retrain_function} as shown in \eqref{eqn:retrain_solution} for all $\hat{c}_{\boldsymbol{d}}$, wherein $p$ represents the set of indices of reconnected channel inputs that were previously removed via pruning (Section~\ref{sec:training_pruning}).
	        This initialisation approach was motivated by the idea that the additional flexibility of the LUTs can be used to compensate for the pruned parts of the network.
	        
            \begin{equation}
                \hat{g}_n{\left(\tilde{\boldsymbol{x}}^{\left(n\right)}\right)} = \hat{w}_1\tilde{x}^{\left(n\right)}_1 + \sum_{r \in p}{\hat{w}_r\tilde{x}^{\left(n\right)}_r}
                \label{eqn:retrain_function}
            \end{equation}

            \begin{equation}
                \hat{c}_{\boldsymbol{d}} = \hat{w}_1 + \sum_{r \in p} d_r \hat{w}_r
                \label{eqn:retrain_solution}
            \end{equation}

	        Once all $\hat{g}_n$ are initialised, our second and final retraining phase is conducted, whereafter the binarised training parameters $c_{\boldsymbol{d}} = \text{sign}{\left(\hat{c}_{\boldsymbol{d}}\right)}$ can be directly interpreted as the configuration mask of each $K$-LUT.
	        
        We elected to follow the network initialisation procedure detailed above rather than training with $K$-LUTs from scratch due to the exponential relationship between $K$ and the number of trainable parameters $\hat{\boldsymbol{c}}$.
	    Training these from the outset, particularly prior to network pruning, would cause both slow convergence and likely overfitting due to the large numbers of local minima in the search space.
	    High-precision training followed by pruning not only ensures fast convergence, it also brings the starting point of $K$-LUT learning closer to global minima, reducing the likelihood of overfitting.
	
	\section{Network Implementation}
	\label{sec:impl}
	
    	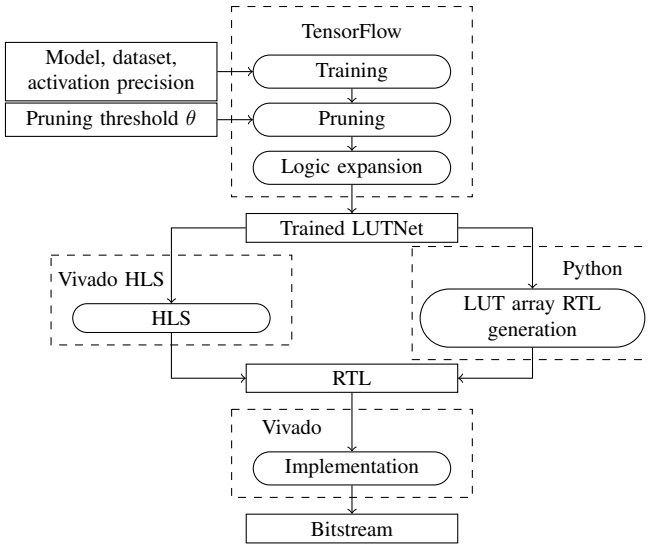
\begin{figure}
    	    \centering
    	    \begin{tikzpicture}[label distance=2mm,decoration={markings,mark= at position 0.5 with{\node[font=\footnotesize] {/};} },scale=0.8, every node/.style={scale=0.8}]

    \tikzstyle{ItemBox} = [rectangle, draw, text centered, minimum height=1em, minimum width=10em, node distance=3cm, text width=9em]
    \tikzstyle{ActionCircle} = [rounded rectangle, draw, text centered, minimum height=1em, minimum width=10em, node distance=3cm, text width=9em]
    \tikzstyle{TextBox} = [rectangle, text centered, minimum height=1em, minimum width=10em, node distance=3cm, text width=9em]
    
    \node[ItemBox] at (0,6) (model) {Model, dataset, activation precision};
    \node[ItemBox] at ($(model)+(0,-0.8)$) (pruning_threshold) {Pruning threshold $\theta$};
    
    \node[ActionCircle] at ($(model)+(4,0)$) (training) {Training};
    \node[ActionCircle] at ($(pruning_threshold)+(4,0)$) (pruning) {Pruning};
    \node[ActionCircle] at ($(pruning)+(0,-0.8)$) (logic_expansion) {Logic expansion};
    
    \node[ItemBox] at ($(logic_expansion)+(0,-1)$) (trained_lutnet) {Trained LUTNet};
    
    \node[ActionCircle] at ($(trained_lutnet)+(-3,-1.5)$) (hls) {HLS};
    \node[ActionCircle] at ($(trained_lutnet)+(3,-1.5)$) (lut_array_rtl_generation) {LUT array RTL generation};
    
    \node[ItemBox] at ($(trained_lutnet)+(0,-2.5)$) (rtl) {RTL};
    
    \node[ActionCircle] at ($(rtl)+(0,-1.5)$) (logic_synthesis) {Implementation};
    
    \node[ItemBox] at ($(logic_synthesis)+(0,-1)$) (bitstream) {Bitstream};
    
    \draw [->] (model.east) -- (training.west);
    \draw [->] (pruning_threshold.east) -- (pruning.west);
    \draw [->] (training.south) -- (pruning.north);
    \draw [->] (pruning.south) -- (logic_expansion.north);
    \draw [->] (logic_expansion.south) -- (trained_lutnet.north);
    
    
    \draw [->] (trained_lutnet.west) -| (hls.north);
    \draw [->] (trained_lutnet.east) -| (lut_array_rtl_generation.north);
    \draw [->] (hls.south) |- (rtl.west);
    \draw [->] (lut_array_rtl_generation.south) |- (rtl.east);
    
    \draw [->] (rtl.south) -- (logic_synthesis.north);
    \draw [->] (logic_synthesis.south) -- (bitstream.north);
    
    \draw [dashed] ($(training.north)+(-2,0.8)$) -- ($(training.north)+(2,0.8)$) -- ($(logic_expansion.south)+(2,-0.2)$) -- ($(logic_expansion.south)+(-2,-0.2)$) -- ($(training.north)+(-2,0.8)$);
    
    \draw [dashed] ($(hls.north)+(-2,0.8)$) -- ($(hls.north)+(2,0.8)$) -- ($(hls.south)+(2,-0.2)$) -- ($(hls.south)+(-2,-0.2)$) -- ($(hls.north)+(-2,0.8)$);
    \draw [dashed] ($(lut_array_rtl_generation.north)+(-2,0.7)$) -- ($(lut_array_rtl_generation.north)+(2,0.7)$) -- ($(lut_array_rtl_generation.south)+(2,-0.2)$) -- ($(lut_array_rtl_generation.south)+(-2,-0.2)$) -- ($(lut_array_rtl_generation.north)+(-2,0.7)$);
    \draw [dashed] ($(logic_synthesis.north)+(-2,0.7)$) -- ($(logic_synthesis.north)+(2,0.7)$) -- ($(logic_synthesis.south)+(2,-0.2)$) -- ($(logic_synthesis.south)+(-2,-0.2)$) -- ($(logic_synthesis.north)+(-2,0.7)$);
    
    \node[TextBox] at ($(training.north)+(0,0.4)$) {TensorFlow};
    \node[TextBox] at ($(hls.north)+(-1,0.4)$) {Vivado HLS};
    \node[TextBox] at ($(lut_array_rtl_generation.north)+(1,0.3)$) {Python};
    \node[TextBox] at ($(logic_synthesis.north)+(-1,0.4)$) {Vivado};

\end{tikzpicture}
        	\caption{LUTNet's fully automated training and FPGA implementation flow.}
        	\label{plot:OVERALL_FLOW}
        \end{figure}
    
        A representation of the overall LUTNet software training and hardware implementation flow is shown in Fig.~\ref{plot:OVERALL_FLOW}. 
        As input, the user provides the desired network model, training dataset, activation precision and the required pruning level to our TensorFlow-based training software, which performs training and pruning.
        Logic expansion is then performed on the chosen layers---also supplied as input---to construct the LUTNet architecture.
        
        We chose to target Xilinx parts for this work, for which two parallel synthesis flows are required in order to convert the trained network into RTL.
        For ease of design and modification, all hardware apart from the inference $K$-LUTs is generated from C templates with Vivado HLS.
        LUT array generation is outsourced to a custom RTL generator written in Python, the output of which is combined with that from Vivado HLS after completion.
        Vivado is then used for implementation.
        
        A separate LUT array generator is required because, as a general-purpose C-to-RTL synthesis tool, Vivado HLS compulsorily performs code transformations and optimisations for the synthesis of efficient RTL.
        Given that LUT configurations are already learnt during training, it is unnecessary---and extremely time-consuming---for such optimisation to be performed on this logic at the C level. 
        Optimisation of RTL LUT arrays at the netlist level during synthesis with Vivado is a lot more efficient, typically taking a few hours---rather than days or weeks---to complete for large designs.
        
	\section{Evaluation}
	\label{sec:eval}
	
	    \subsection{Benchmarks}
	    \label{sec:eval_bench}

            \begin{table*}
        		\centering
        		\caption{
        		    Network architectures for evaluated benchmarks.
        		    Conv\textsubscript{$x,y,z$} denotes a convolutional layer with $x$ outputs, kernel size $y \times y$ and stride $z$.
        		    FConn\textsubscript{$x$} is a fully connected layer with $x$ outputs.
        		    MaxPool\textsubscript{$x$} is an $x \times x$ maximum-pooling layer, and BatchNorm and SoftMax are batch normalisation and normalised exponential layers, respectively.
        		    Layers in bold were fully unrolled and, for LUTNet, feature $K$-LUT inference operators.
                }
    		    \begin{tabular}{ccc}
					\toprule
					Dataset			 		& Model	& Network architecture    \\
					\midrule
					\multirow{2}{*}{MNIST}  & \multirow{2}{*}{LFC}  & FConn\textsubscript{256}, BatchNorm, \textbf{FConn\textsubscript{256}}, BatchNorm, \textbf{FConn\textsubscript{256}}, BatchNorm, \textbf{FConn\textsubscript{256}}, BatchNorm, \textbf{FConn\textsubscript{10}}, 	\\
											&		& BatchNorm, SoftMax \\
					\midrule
											&   	& Conv\textsubscript{64, 3, 1}, BatchNorm, Conv\textsubscript{64, 3, 1}, BatchNorm, MaxPool\textsubscript{2}, Conv\textsubscript{128, 3, 1}, BatchNorm,	Conv\textsubscript{128, 3, 1}, BatchNorm, \\
					SVHN \& CIFAR-10		& CNV   & MaxPool\textsubscript{2}, Conv\textsubscript{256, 3, 1}, BatchNorm, \textbf{Conv\textsubscript{256, 3, 1}}, BatchNorm, FConn\textsubscript{512}, BatchNorm, FConn\textsubscript{512}, BatchNorm, 	\\
											&		& FConn\textsubscript{10}, BatchNorm, SoftMax	\\
					\midrule
											&		& Conv\textsubscript{96, 11, 4}, BatchNorm, MaxPool\textsubscript{3}, Conv\textsubscript{256, 5, 1}, BatchNorm, MaxPool\textsubscript{3}, Conv\textsubscript{384, 3, 1}, BatchNorm, Conv\textsubscript{384, 3, 1}, 	\\
					ImageNet				& AlexNet   & BatchNorm, \textbf{Conv\textsubscript{256, 3, 1}}, BatchNorm, MaxPool\textsubscript{3}, FConn\textsubscript{4096}, BatchNorm, FConn\textsubscript{4096}, BatchNorm, FConn\textsubscript{1000}, 	\\
											&		& BatchNorm, SoftMax   \\
					\bottomrule
				\end{tabular}
    			\label{tab:model_info}
    		\end{table*}

            For evaluation, we implemented end-to-end dataflow engines for the DNN models shown in Table~\ref{tab:model_info}, using them to classify the listed datasets.
            Our primary baseline was the state-of-the-art BNN architecture, ReBNet~\cite{BNN_CNN_REBNET_FCCM}.
            All hardware implementations targetted the Xilinx Kintex UltraScale XCKU115 FPGA and met timing at 200~MHz.
            
            In order to demonstrate the capabilities of specialised LUTs, we unrolled a subset of each network such that each node within that subset mapped to a distinct compute unit.
            We chose to unroll by layer, unrolling as many layers as the target device could accommodate and implementing those following the LUTNet approach.
            Those selected for unrolling are marked in bold in Table~\ref{tab:model_info}.
            For fairness of comparison, BNN architectures (chiefly ReBNet) used as baselines had the same layers unrolled, and fine-grained pruning was performed identically to that carried out for LUTNet on those layers.
            The remaining layers were left time-multiplexed, with identical folding factors to those used for ReBNet's evaluation.
    
        \subsection{Training Particulars}
        \label{sec:eval_training}
        
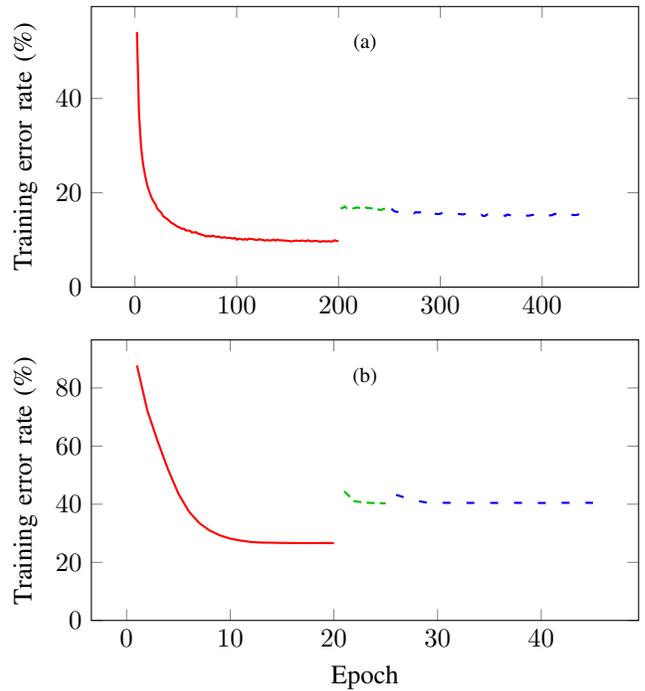
\begin{figure}
                \begin{tikzpicture}
    
    \begin{groupplot} [
		width=\columnwidth,
		height=0.6\columnwidth,
		group style={group size=1 by 2, xlabels at=edge bottom, vertical sep=2em},
		ymin=0,
		xlabel near ticks,
		xlabel={Epoch},
        ylabel near ticks,
        ylabel={Training error rate (\%)}
	]
        
        \nextgroupplot
            
            \addplot [thick, red] table [y=training_error, x=epoch] {data/cifar_training_curve/train.txt}; \label{plt:cifar_training_curve_train}
            \addplot [thick, black!25!green, densely dashed] table [y=training_error, x=epoch] {data/cifar_training_curve/prune.txt}; \label{plt:cifar_training_curve_prune}
            \addplot [thick, blue, loosely dashed] table [y=training_error, x=epoch] {data/cifar_training_curve/retrain.txt}; \label{plt:cifar_training_curve_retrain}
            
            \node [text width=1em, anchor=north] at (axis description cs:0.5, 1) {\subcaption{\label{plot:cifar_training_curve}}};
            
        \nextgroupplot
        
            \addplot [thick, red] table [y=training_error, x=epoch] {data/imagenet_training_curve/train.txt}; \label{plt:imagenet_training_curve_train}
            \addplot [thick, black!25!green, densely dashed] table [y=training_error, x=epoch] {data/imagenet_training_curve/prune.txt}; \label{plt:imagenet_training_curve_prune}
            \addplot [thick, blue, loosely dashed] table [y=training_error, x=epoch] {data/imagenet_training_curve/retrain.txt}; \label{plt:imagenet_training_curve_retrain}
            
            \node [text width=1em, anchor=north] at (axis description cs:0.5, 1) {\subcaption{\label{plot:imagenet_training_curve}}};
    
    \end{groupplot}

\end{tikzpicture}

            	\caption{
            	    Training losses for (\subref{plot:cifar_training_curve}) the CNV network classifying the CIFAR-10 dataset and (\subref{plot:imagenet_training_curve}) AlexNet classifying ImageNet.
            	    Curves represent high-precision training~(\ref{plt:cifar_training_curve_train}), high-precision post-pruning retraining~(\ref{plt:cifar_training_curve_prune}) and post-logic expansion retraining with binarised forward propagation~(\ref{plt:cifar_training_curve_retrain}).
                }
            	\label{plot:training_curves}
            \end{figure}

            For our simpler datasets (MNIST, SVHN and CIFAR-10), we performed the training, post-pruning retraining and post-logic expansion retraining described in Section~\ref{sec:training} for 200, 50 and 200 epochs, respectively. 
            For the more complex ImageNet dataset, these were performed for 20, 5 and 20 epochs instead.
            These periods were selected from our observations during training, the loss curves for which are shown in Fig.~\ref{plot:cifar_training_curve}, demonstrating saturation at or before these epochs.
            Non-LUTNet implementations were identically trained, but the logic expansion phase (Section~\ref{sec:training_expansion}) was not performed.
            All training phases were executed in TensorFlow and accelerated using four Nvidia GTX~1080~Ti GPUs.
            
        \subsection{Area Efficiency}
        \label{sec:eval_area_accuracy}
            
            When evaluating our implementations, we were primarily interested in \emph{logic density}, which we define as the number of LUTs required to construct a network able to achieve a particular test accuracy for a given dataset.
            The fewer LUTs needed to reach the same accuracy, the higher the density and thus the more efficient the implementation.

\begin{figure}
                \centering
                \begin{tikzpicture}

    \begin{axis}[
		width=\columnwidth,
		height=\columnwidth,
		xlabel near ticks,
		xlabel={Area occupancy (LUTs)},
		ylabel near ticks,
		ylabel={Test error rate (\%)},
        xmin=100000,
        xmax=550000,
        error bars/y dir      = both,
        error bars/y explicit = true,
    ]
        \addplot [thick, only marks, mark=x, mark options={scale=1.5, color=red}] table [y=RPerr, x=RPLUT, y error plus=RerrbarU, y error minus=RerrbarL] {data/err_area.txt}; \label{plt:tradeoff_rebnet}
        \addplot [thick, only marks, mark=o, mark options={scale=1.5, color=black!25!green}] table [y=2Lerr, x=2LLUT, y error plus=2errbarU, y error minus=2errbarL] {data/err_area.txt}; \label{plt:tradeoff_2lutnet}
        \addplot [thick, only marks, mark=+, mark options={scale=1.5, color=blue}] table [y=4Lerr, x=4LLUT, y error plus=4errbarU, y error minus=4errbarL] {data/err_area.txt}; \label{plt:tradeoff_4lutnet}
        \addplot [thick, only marks, mark=*, mark options={scale=1.5, color=black!25!cyan}] table [y=6Lerr, x=6LLUT, y error plus=6errbarU, y error minus=6errbarL] {data/err_area.txt}; \label{plt:tradeoff_6lutnet}
        \addplot [thick, dashed] coordinates {(100000,15.55) (550000,15.55)};
	\end{axis}

\end{tikzpicture}



            	\caption{
            	    Area-accuracy tradeoff for pruned ReBNet~\cite{BNN_CNN_REBNET_FCCM}~(\ref{plt:tradeoff_rebnet}), 2-LUTNet~(\ref{plt:tradeoff_2lutnet}), 4-LUTNet~(\ref{plt:tradeoff_4lutnet}) and 6-LUTNet~(\ref{plt:tradeoff_6lutnet}) with the CNV network and CIFAR-10 dataset.
            	    Each point is representative of a distinct pruning threshold.
            	    The dashed line shows the baseline accuracy for unpruned ReBNet (660196 LUTs).
                }
            	\label{plot:AREA_LUT_TRADEOFF}
            \end{figure}
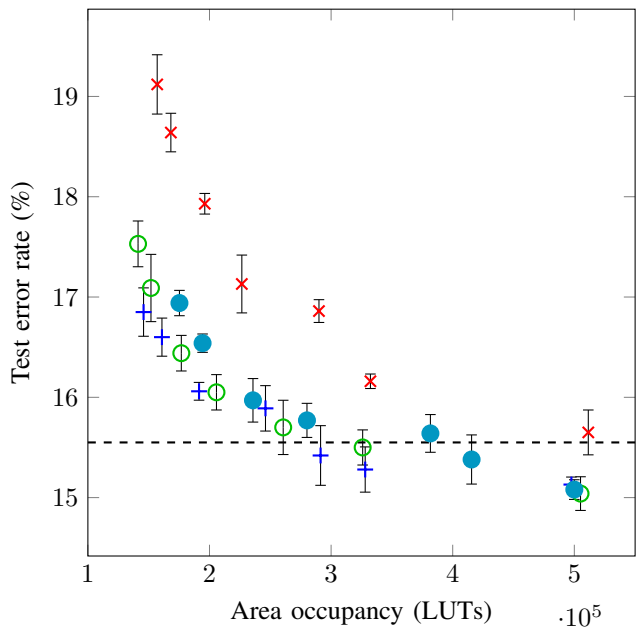

            Fig.~\ref{plot:AREA_LUT_TRADEOFF} shows the achieved whole-network area \emph{vs} test accuracy points for ReBNet and LUTNet implementations, each pruned to various densities (proportion of remaining pre-pruning parameters) via the tuning of pruning threshold $\theta$, for CNV classifying CIFAR-10.
            Each point marks the mean of five independent training runs, with an error bar indicating its 90\% confidence interval.
            LUTNet implementations used 2-, 4- and 6-LUT inference operators.
            For reference, the mean test error rate of ReBNet without pruning---again averaged over five training runs---is also shown.
            From this data, one can clearly observe that while the error rate increases as more aggressive pruning is applied, LUTNet demonstrates greater robustness to that pruning than ReBNet through its increased logic density.
            That several LUTNet points achieve greater test accuracy than the unpruned baseline speaks to LUTNet's increased expressiveness.
            For example, despite having a significantly lower (2.27$\times$) area requirement, our 91.1\%-pruned 4-LUTNet implementation achieved an accuracy 0.590~pp above that of the ReBNet implementation without pruning.
            
            It is interesting to note from Fig.~\ref{plot:AREA_LUT_TRADEOFF} that 6-LUTNet implementations tended to achieve lower logic densities than those of 4- and sometimes even 2-LUTNet.
            To understand why this is the case, we must consider area and accuracy separately.
            
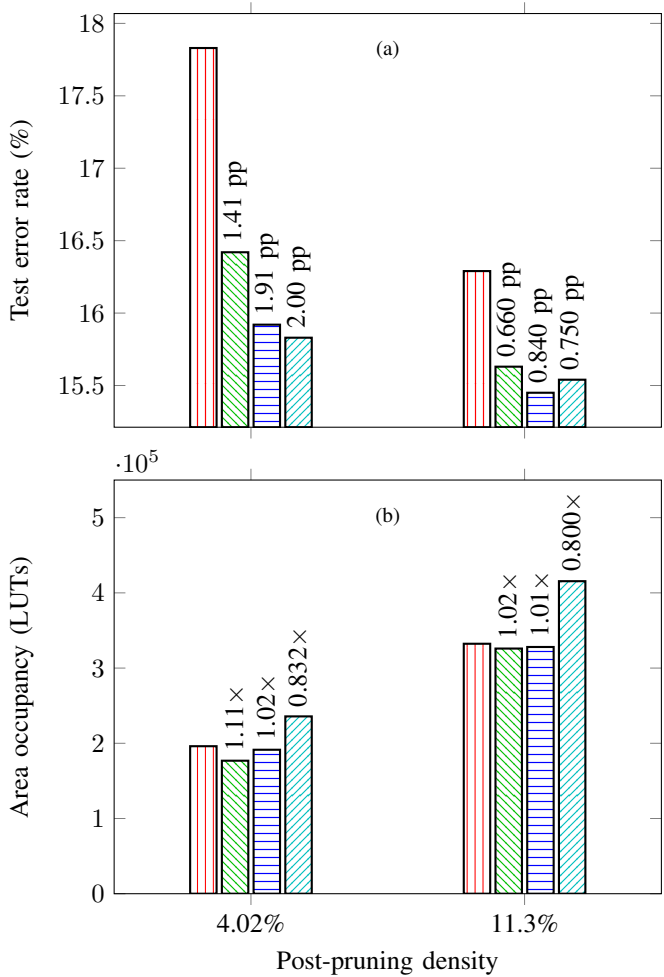
\begin{figure}
                \begin{tikzpicture}
    
    \begin{groupplot} [
		ybar,
		width=\columnwidth,
		height=0.8\columnwidth,
		group style={group size=1 by 2, xlabels at=edge bottom, xticklabels at=edge bottom, vertical sep=2em},
		xlabel near ticks,
		xlabel={Post-pruning density},
		xtick={1, 2},
        xticklabels={4.02\%, 11.3\%},
        enlarge x limits=0.5,
        legend image code/.code={
            \draw[#1, bar width=6pt, yshift=-0.3em] plot coordinates {(0cm,0.8em)};
        }
	]
        
        \nextgroupplot [
    		ylabel={Test error rate (\%)}
        ]
            \addplot[thick, pattern=vertical lines, pattern color=red] coordinates {(1, 17.83) (2, 16.29)};
            \label{plt:trn_err_rebnet_low}
            \addplot[thick, pattern=north west lines, pattern color=black!25!green] coordinates {(1, 16.42) (2, 15.63)}
            node [pos=0, xshift=-0.6*\pgfkeysvalueof{/pgf/bar width}, rotate=90, anchor=west] {1.41~pp}
            node [pos=1, xshift=-0.6*\pgfkeysvalueof{/pgf/bar width}, rotate=90, anchor=west] {0.660~pp};
            \label{plt:trn_err_2lutnet_low}
            \addplot[thick, pattern=horizontal lines, pattern color=blue] coordinates {(1, 15.92) (2, 15.45)}
            node [pos=0, xshift=0.6*\pgfkeysvalueof{/pgf/bar width}, rotate=90, anchor=west] {1.91~pp}
            node [pos=1, xshift=0.6*\pgfkeysvalueof{/pgf/bar width}, rotate=90, anchor=west] {0.840~pp};
            \label{plt:trn_err_4lutnet_low}
            \addplot[thick, pattern=north east lines, pattern color=black!25!cyan] coordinates {(1, 15.83) (2, 15.54)}
            node [pos=0, xshift=1.8*\pgfkeysvalueof{/pgf/bar width}, rotate=90, anchor=west] {2.00~pp}
            node [pos=1, xshift=1.8*\pgfkeysvalueof{/pgf/bar width}, rotate=90, anchor=west] {0.750~pp};
            \label{plt:trn_err_6lutnet_low}
            
        \node [text width=1em, anchor=north] at (axis description cs:0.5, 1) {\subcaption{\label{plot:TRAINING_LOSS}}};
        
        \nextgroupplot [
            ymin=0,
            ymax= 550000,
    		ylabel={Area occupancy (LUTs)}
        ]
        
            \addplot[thick, pattern=vertical lines, pattern color=red] coordinates {(1, 196117) (2, 332339)};
            \label{plt:util_rebnet}
            \addplot[thick, pattern=north west lines, pattern color=black!25!green] coordinates {(1, 176833) (2, 325959)}
            node [pos=0, xshift=-0.6*\pgfkeysvalueof{/pgf/bar width}, rotate=90, anchor=west] {1.11$\times$}
            node [pos=1, xshift=-0.6*\pgfkeysvalueof{/pgf/bar width}, rotate=90, anchor=west] {1.02$\times$};
            \label{plt:util_2lutnet}
            \addplot[thick, pattern=horizontal lines, pattern color=blue] coordinates {(1, 191405) (2, 328041)}
            node [pos=0, xshift=0.6*\pgfkeysvalueof{/pgf/bar width}, rotate=90, anchor=west] {1.02$\times$}
            node [pos=1, xshift=0.6*\pgfkeysvalueof{/pgf/bar width}, rotate=90, anchor=west] {1.01$\times$};
            \label{plt:util_4lutnet}
            \addplot[thick, pattern=north east lines, pattern color=black!25!cyan] coordinates {(1, 235757) (2, 415529)}
            node [pos=0, xshift=1.8*\pgfkeysvalueof{/pgf/bar width}, rotate=90, anchor=west] {0.832$\times$}
            node [pos=1, xshift=1.8*\pgfkeysvalueof{/pgf/bar width}, rotate=90, anchor=west] {0.800$\times$};
            \label{plt:util_6lutnet}
            
        \node [text width=1em, anchor=north] at (axis description cs:0.5, 1) {\subcaption{\label{plot:UTIL_AT_SAME_PRUNING}}};
    
    \end{groupplot}

\end{tikzpicture}
	
            	\caption{
            	(\subref{plot:TRAINING_LOSS}) Accuracy and (\subref{plot:UTIL_AT_SAME_PRUNING}) area for ReBNet~\cite{BNN_CNN_REBNET_FCCM}~(\ref{plt:trn_err_rebnet_low}), 2-LUTNet~(\ref{plt:trn_err_2lutnet_low}), 4-LUTNet~(\ref{plt:trn_err_4lutnet_low}) and 6-LUTNet~(\ref{plt:trn_err_6lutnet_low}) with CNV, pruned to two densities, classifying CIFAR-10.
            	Annotations denote decreases \emph{vs} ReBNet.
            	}
            	\label{plot:area_accuracy_pruned}
            \end{figure}

            Fig.~\ref{plot:TRAINING_LOSS} shows the test accuracies of the same implementations---ReBNet, 2-, 4- and 6-LUTNet---for the same network and dataset---CNV and CIFAR-10---pruned to two densities: 4.02\% and 11.3\%.
            These densities were selected for comparison since they represent a wide spread over those found to achieve accuracies reasonably close ($\pm$2.00~pp) to ReBNet when unpruned.
            Of particular pertinence is the difference in accuracy spread between the two: those at 11.3\% are much tighter than their 4.02\% parallels.
            These diminishing accuracy returns when adding LUT inputs at higher densities point to complexity saturation.
            
            Turning now to area, Fig.~\ref{plot:UTIL_AT_SAME_PRUNING} shows the LUT requirements of the same implementations.
            While $K$-LUTNet designs for any $K$ with equal density contain the same number of logical LUTs, this does not mean that they consume the same number of physical LUTs.
            The LUTs actually present in our target device are 6-LUTs, each capable of implementing either a single logical 6-LUT or two logical $K$-LUTs with at least five (for 5-LUTs), three (4-LUTs) or one (3-LUTs) shared inputs.
            1- and 2-LUTs are not required to share any inputs; two of these can always be packed together.
            For 2- and 4-LUTNet, in which each inference operator uses fewer than five inputs, Vivado can often (for 4-LUTNet) or always (2-LUTNet) pack two logical $K$-LUTs into each physical 6-LUT, resulting in high logic density.
            Training-induced simplifications, \emph{e.g.} inputs treated as don't-cares that are removed during synthesis, also lead to higher probabilities of additional packing when smaller logical LUTs are used.
            These optimisation phenomena are rarely seen for 6-LUTNet, hence its significantly higher area requirements at equal density.
            
            When moving from 4- to 6-LUTs at the higher density, despite the $>$20\% increase in physical LUTs, no accuracy benefit was obtained.
            In fact, 6-LUTNet's accuracy actually fell $\sim$0.1~pp below that of 4-LUTNet's as a result of overfitting.
            Due to this, as was shown in Fig.~\ref{plot:AREA_LUT_TRADEOFF}, 4-LUTNet almost always achieves a better area-accuracy tradeoff than 6-LUTNet.
            
\begin{figure}
                \centering
                \begin{tikzpicture}
    
    \begin{axis}[ybar,
		width=\columnwidth,
		height=0.8\columnwidth,
        ymin=0,
        ylabel near ticks,
		ylabel={Area occupancy (LUTs)},
		xtick={1, 2, 3, 4},
        xticklabels={\shortstack{MNIST\\(LFC)}, \shortstack{SVHN\\(CNV)}, \shortstack{CIFAR-10\\(CNV)}, \shortstack{ImageNet\\(AlexNet)}},
        enlarge x limits=0.3,
        x tick label style={rotate=60, anchor=east},
        xlabel near ticks,
        xlabel={Dataset (network)},
        legend image code/.code={
            \draw[#1, bar width=6pt, yshift=-0.3em] plot coordinates {(0cm,0.8em)};
        }
    ]
        \addplot [thick, pattern=vertical lines, pattern color=red] coordinates {(1, 48102) (2, 379403) (3, 511494) (4, 941768)};
        \label{plt:area_comp_rebnet_wup};
        \addplot [thick, pattern=north west lines, pattern color=blue] coordinates {(1, 58192) (2, 154814) (3, 246044) (4, 496106)}
        node [pos=0, xshift=0.6*\pgfkeysvalueof{/pgf/bar width}, rotate=90, anchor=west] {0.827$\times$}
        node [pos=0.333, xshift=0.6*\pgfkeysvalueof{/pgf/bar width}, rotate=90, anchor=west] {2.45$\times$}
        node [pos=0.667, xshift=0.6*\pgfkeysvalueof{/pgf/bar width}, rotate=90, anchor=west] {2.08$\times$}
        node [pos=1, xshift=0.6*\pgfkeysvalueof{/pgf/bar width}, rotate=90, anchor=west] {1.90$\times$};
        \label{plt:area_comp_4lutnet};
    \end{axis}

\end{tikzpicture}
	
            	\caption{
            	    Area occupancy for ReBNet~\cite{BNN_CNN_REBNET_FCCM}~(\ref{plt:area_comp_rebnet_wup}) and 4-LUTNet~(\ref{plt:area_comp_4lutnet}) with various models and datasets.
            	    Via pruning, each implementation's test accuracy was kept within $\pm$0.300~pp of that of the unpruned ReBNet baseline's.
            	    Annotations show the area decrease in each case.
                }
            	\label{plot:area_compression}
            \end{figure}
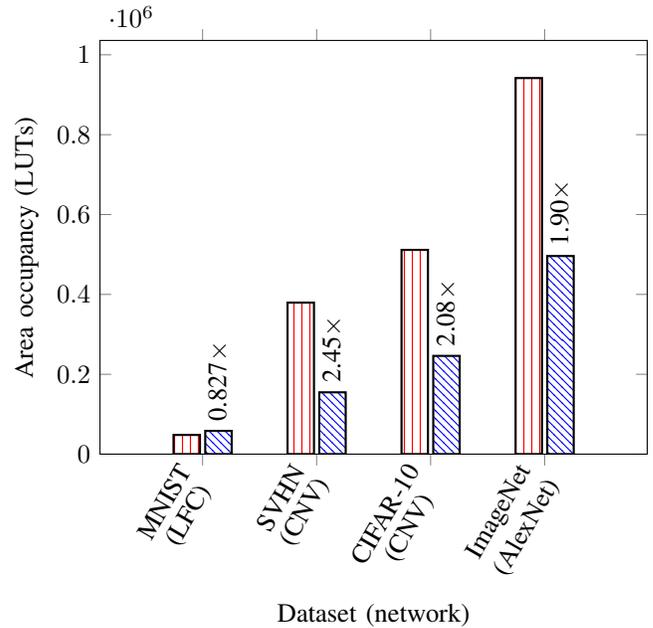

            As was noted in Section~\ref{sec:eval_bench}, we also benchmarked LUTNet on other popular datasets and models: MNIST (on LFC), SVHN (on CNV) and ImageNet (on AlexNet).
            Fig.~\ref{plot:area_compression} shows the LUT requirements of each of these model-dataset combinations when implemented using both the ReBNet and LUTNet inference architectures.
            The same layers for all pairs of implementations were fully unrolled and pruned, with the pruning threshold tuned to achieve an accuracy degradation no more than $\pm$0.300~pp \emph{vs} ReBNet's without pruning.
            
            For CNV and AlexNet, our use of arbitrary inference operators sees area reductions of around 2$\times$.
            For the classification of SVHN, the CNV network used can be pruned more heavily than for CIFAR-10, hence the greater area saving for that dataset.
            For LFC classifying MNIST, however, more LUTs were consumed by LUTNet than its pruned ReBNet counterpart. 
            While each of CNV's hidden layers has 2304 inputs per channel, LFC's channels each have only 256 inputs, presenting less opportunity for area reduction through popcount simplification.
            In this case, LUTNet's post-pruning LUT savings through popcount tree thinning were unable to make up for the inference operator LUT incursion.
        
        \subsection{Area Breakdown}
        \label{sec:eval_breakdown}
            
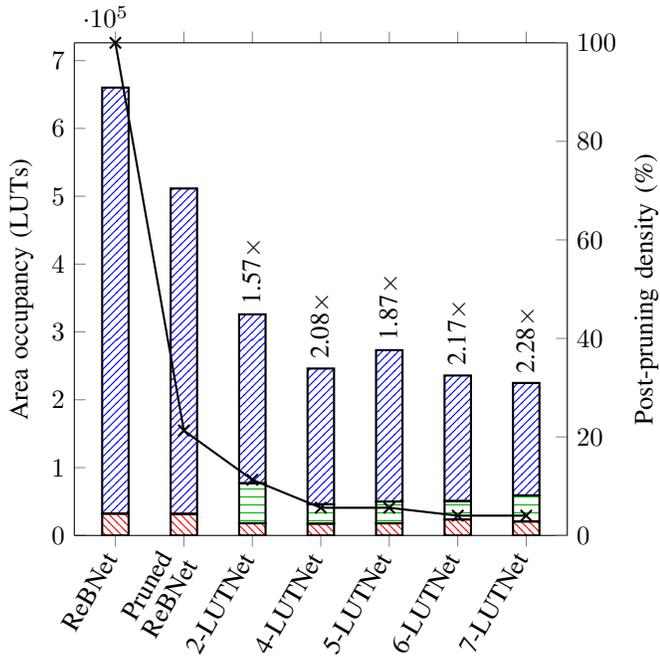
\begin{figure}
                \centering
                \begin{tikzpicture}
    
    \begin{axis}[
        ybar stacked,
        height=0.74\columnwidth,
        width=0.74\columnwidth,
		scale only axis,
		axis y line*=left,
        ymin=0,
        ylabel near ticks,
		ylabel={Area occupancy (LUTs)},
        xtick=data,
        xticklabels from table={data/util_detail.txt}{Name},
        x tick label style={rotate=60, anchor=east},
        xtick align=outside,
        legend image code/.code={
            \draw[#1, bar width=6pt, yshift=-0.3em] plot coordinates {(0cm,0.8em)};
        }
    ]
        \addplot [thick, pattern=north west lines, pattern color=red] table [x=id, y=otherlayers] {data/util_detail.txt};
        \label{plt:breakdown_otherlayers}
        \addplot [thick, pattern=horizontal lines, pattern color=black!25!green] table [x=id, y=ops] {data/util_detail.txt};
        \label{plt:breakdown_ops}
        \addplot [thick, pattern=north east lines, pattern color=blue] table [x=id, y=popcount] {data/util_detail.txt}
        node [pos=0.333, rotate=90, anchor=west] {1.57$\times$}
        node [pos=0.5, rotate=90, anchor=west] {2.08$\times$}
        node [pos=0.667, rotate=90, anchor=west] {1.87$\times$}
        node [pos=0.833, rotate=90, anchor=west] {2.17$\times$}
        node [pos=1, rotate=90, anchor=west] {2.28$\times$};
        \label{plt:breakdown_popcount}
    \end{axis}

    \begin{axis}[
        height=0.74\columnwidth,
        width=0.74\columnwidth,
		scale only axis,
        axis y line*=right,
        ymin=0,
        ymax=100,
        ylabel near ticks,
        ylabel={Post-pruning density (\%)},
        axis x line=none,
    ]
        \addplot [thick, mark=x, mark options={scale=1.5}] table [x=id, y=Density] {data/util_detail.txt};
        \label{plt:breakdown_density}
	\end{axis}

\end{tikzpicture}
	
            	\caption{
            	    LUT use breakdown, presented in terms of popcount operators~(\ref{plt:breakdown_popcount}), inference operators~(\ref{plt:breakdown_ops}) and other layers~(\ref{plt:breakdown_otherlayers}), for CNV implementations.
            	    Each implementation's test accuracy was within $\pm$0.300~pp of that of the unpruned ReBNet baseline's~\cite{BNN_CNN_REBNET_FCCM}.
            	    Points~(\ref{plt:breakdown_density}) show post-pruning densities.
            	    Annotations show decreases \emph{vs} ReBNet with pruning.
                }
            	\label{plot:AREA_SAVINGS}
            \end{figure}

            As a crude method of verifying the source of LUTNet's area savings, we disabled design hierarchy optimisation in Vivado, preventing the synthesis engine from flattening across modules.
            By taking a slice of implementations shown in Fig.~\ref{plot:AREA_LUT_TRADEOFF} at the unpruned ReBNet test error rate (84.5\%) $\pm$0.300~pp, we obtained pruned ReBNet and 2-, 4-, 5-, 6- and 7-LUTNet implementations for CNV all of comparable CIFAR-10 test accuracy.
            Fig.~\ref{plot:AREA_SAVINGS} shows the LUT requirements for each of these, with area split into that required by popcount operators, inference operators and everything else.
            The overall height of each bar is the whole design's area occupancy with hierarchy optimisation \emph{enabled}, but the height of each stacked bar is relative to the proportional area obtained with hierarchy optimisation \emph{disabled}.
            We emphasise that these relative area data are not particularly meaningful, however this was the best we could do without significant manual tool intervention.
            
            Generally, as more inputs are used per logical LUT, we can see that physical LUT requirements decrease, highlighting $K$-LUTNet's increasing logic density with $K$.
            Also shown in Fig.~\ref{plot:AREA_SAVINGS} is each implementation's post-pruning density.
            From the breakdowns, it can be seen that the number of physical LUTs required for popcount operators drops dramatically with density.
            More aggressive pruning reduces the number of branches in each popcount tree which, when unrolled, consume the majority of the target device's area.
            
            As was pointed out in Section~\ref{sec:intro}, due to following a traditional BNN inference paradigm, ReBNet implementations---whether pruned or not---require zero LUTs for the realisation of their inference operators since, when unrolled, XNORs become free-to-implement buffers and inverters.
            For LUTNet, this is not the case: physical LUTs are consumed by our logical $K$-LUTs.
            As shown in Fig.~\ref{plot:AREA_SAVINGS}, however, this is more than outweighed by significant popcount area reduction.
            This confirms the statement made in Section~\ref{sec:intro} regarding $\tilde{N} \ll N$.
            
            Between 2- and 6-LUTNet, we can observe a general trend of decreasing inference operator LUT requirements with density.
            Looking more closely, some more interesting features emerge.
            The jump in total area between 4- and 5-LUTNet can be attributed to two factors: lack of density reduction and decreased opportunity for LUT sharing.
            Here, the increased expressiveness of 5-LUTs was not significant enough to enable increased pruning while remaining within the required accuracy bound.
            On top of this, the logical-to-physical LUT packing effects discussed in Section~\ref{sec:eval_area_accuracy} were marked, pushing both inference operator and total LUT requirements for 5-LUTNet above those for 4-LUTNet.
            Thereafter, although increasing numbers of physical LUTs were occupied by the 6- and 7-LUTNet implementations, decreases in density facilitated through increased network complexity caused more-than-compensatory popcount area reductions.

        \subsection{Energy Efficiency}
        \label{sec:energy_eff}

\begin{figure}
        	\begin{tikzpicture}
    
    \begin{axis}[
        ybar stacked,
        height=0.74\columnwidth,
        width=0.74\columnwidth,
		scale only axis,
		axis y line*=left,
        ymin=0,
        ylabel near ticks,
		ylabel={Power consumption (W)},
        xtick=data,
        xticklabels from table={data/energy_efficiency.txt}{Name},
        x tick label style={rotate=60, anchor=east},
        xtick align=outside,
        legend image code/.code={
            \draw[#1, bar width=6pt, yshift=-0.3em] plot coordinates {(0cm,0.8em)};
        }
    ]
        \addplot [thick, pattern=north west lines, pattern color=red] table [x=id, y=static] {data/energy_efficiency.txt};
        \label{plt:static_power}
        \addplot [thick, pattern=north east lines, pattern color=blue] table [x=id, y=dynamic] {data/energy_efficiency.txt}
        node [pos=0.333, rotate=90, anchor=west] {4.43$\times$}
        node [pos=0.5, rotate=90, anchor=west] {5.85$\times$}
        node [pos=0.667, rotate=90, anchor=west] {6.66$\times$}
        node [pos=0.833, rotate=90, anchor=west] {6.37$\times$}
        node [pos=1, rotate=90, anchor=west] {6.11$\times$};
        \label{plt:dynamic_power}
    \end{axis}

    \begin{axis}[
        height=0.74\columnwidth,
        width=0.74\columnwidth,
		scale only axis,
        axis y line*=right,
        ymin=0,
        ymax=100,
        ylabel near ticks,
        ylabel={Post-pruning density (\%)},
        axis x line=none,
    ]
        \addplot [thick, mark=x, mark options={scale=1.5}] table [x=id, y=Density] {data/energy_efficiency.txt};
        \label{plt:energy_density}
	\end{axis}

\end{tikzpicture}
	
        	\caption{
                Implementation power consumption estimates, broken into static~(\ref{plt:static_power}) and dynamic~(\ref{plt:dynamic_power}) components, for the same CNV implementations used in Fig.~\ref{plot:AREA_SAVINGS}.
                Points~(\ref{plt:breakdown_density}) show post-pruning densities.
                Annotations show decreases \emph{vs} ReBNet with pruning.
            }
        	\label{plot:energy_efficiency}
        \end{figure}
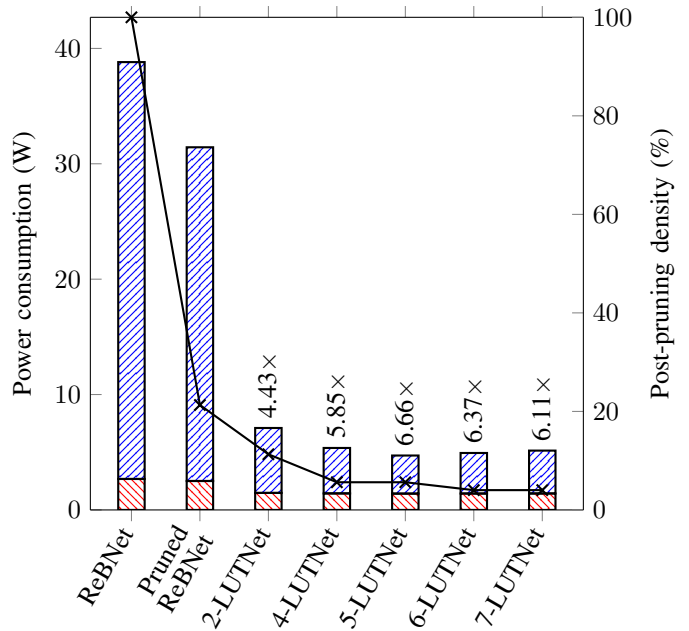

        We estimated LUTNet's energy efficiency using the Xilinx Power Analyzer (XPA) tool with default settings: vectorless mode (that not requiring specific input stimuli) and 12.5\% primary input switching probability.
        The resultant power estimates, for the same implementations captured in Fig.~\ref{plot:AREA_SAVINGS}, are shown in Fig.~\ref{plot:energy_efficiency}.
        All were obtained post-placement and \-/routing.
        Power consumption is equivalent to energy efficiency here since all implementations have identical throughput.
        While we acknowledge that vectorless power estimates are not particularly accurate---typically around $\pm$10--20\% from measured values~\cite{KAPOW}---they are sufficiently so for our purposes.
        
        Since dynamic power consumption is directly related to area occupancy, Figs~\ref{plot:AREA_SAVINGS} and \ref{plot:energy_efficiency} show similar trends.
        Most of the fully unrolled networks' area consumption is attributable to popcount adder trees, whose carry chains are dominant with respect to switching activity.
        Popcount branch pruning shortens the chains, more than proportionately lowering their switching rates and thereby causing the large dynamic power drop.
        The reduction in static power between the ReBNet and LUTNet implementations can also be linked to area, although indirectly.
        Between Pruned ReBNet and 2-LUTNet there was a drop in estimated junction temperature from 60.1\textdegree~to 31.3\textdegree, leading to reduced leakage current and therefore static power draw.
        Such temperature decreases are also useful since they limit ageing, thereby increasing device lifetime~\cite{DEGRADATION}.
        Overall, we can conclude that LUTNet's significant area reductions result in even greater energy efficiency improvements.
        
        \subsection{Training Efficiency}
        \label{sec:eval_training_eff}
        
\begin{figure}
                \begin{tikzpicture}
    
    \begin{axis}[
        ybar,
		width=\columnwidth,
		height=0.8\columnwidth,
        ymin=0,
        ymax=180,
        ylabel near ticks,
		ylabel={Training time per epoch (s)},
        xtick=data,
        xticklabels from table={data/training_speed.txt}{Name},
        x tick label style={rotate=60, anchor=east}
    ]
        \addplot [thick] table [x expr=\coordindex, y={training_time}] {data/training_speed.txt}
        node [pos=0.167, rotate=90, anchor=west] {1.00$\times$}
        node [pos=0.333, rotate=90, anchor=west] {1.03$\times$}
        node [pos=0.5, rotate=90, anchor=west] {1.03$\times$}
        node [pos=0.667, rotate=90, anchor=west] {1.67$\times$}
        node [pos=0.833, rotate=90, anchor=west] {2.24$\times$}
        node [pos=1, rotate=90, anchor=west] {4.33$\times$};
    \end{axis}

\end{tikzpicture}
	
            	\caption{
            	    Training efficiency of ReBNet~\cite{BNN_CNN_REBNET_FCCM} and LUTNet implementations for CNV classifying CIFAR-10.
            	    Annotations show increases over ReBNet.
                }
            	\label{plot:TRAINING_SPEED}
            \end{figure}
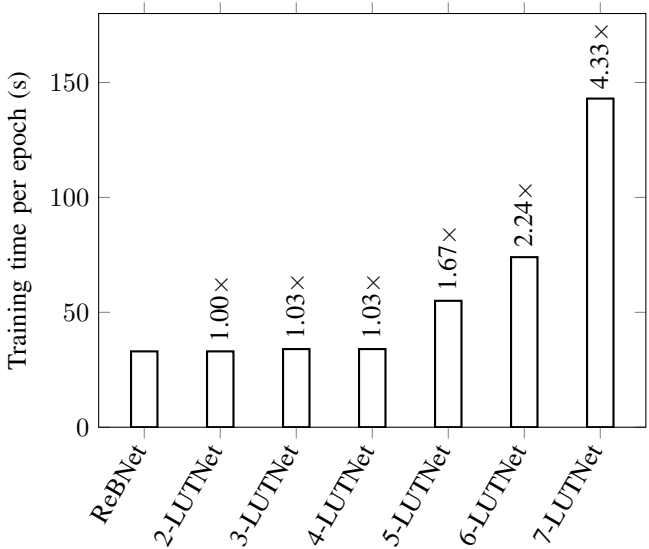

            Each of LUTNet's inference $K$-LUTs consists of 2\textsuperscript{$K$} weights: 2$\times$ more than that for $\left(K-1\right)$-LUTNet.
            Consequently, the number of training operations required per epoch increases exponentially with $K$.
            This does not necessarily translate to exponentially increasing training times over XNOR-based BNNs, however, since, as pointed out by Jouppi \emph{et al.}, the majority of DNN training accelerators' speed is bounded by memory bandwidth, not compute power~\cite{FXP_CNN_TPU}.
            This is evident from Fig.~\ref{plot:TRAINING_SPEED}, which shows the per-epoch training times of ReBNet and 2-, 3-, 4-, 5-, 6- and 7-LUTNet implementations for CNV with CIFAR-10.
            Implementations from ReBNet to 4-LUTNet all have approximately the same training rate, despite the number of weights increasing by up to 16$\times$.
            The training time did not increase because, for all of these implementations, progress was bottlenecked by high-precision activation transfer to and from GPU RAM.
            Increases of significance were seen for 5-LUTNet and beyond, for which the number of multiply-accumulate operations performed per activation transferred rose enough for the former to dominate.
            
	\section{Limitations}
	\label{sec:limit}
        
        While LUTNet implementations typically reach significantly higher logic density than XNOR-based BNNs, our proposal's greatest current limitation is that its inference $K$-LUTs cannot be time-multiplexed.
        Consequently, DNN hardware---in this paper, always complete layers---implemented following the LUTNet approach must be fully unrolled.
        While this may be acceptable in, for example, cloud deployments where throughput and energy efficiency are of paramount importance~\cite{ASAP}, it nevertheless limits the scalability of our proposal.
        
        Time-multiplexing could be introduced in several ways.
        By adding a level of multiplexers prior to $K$-LUTs used as we propose herein, \emph{i.e.} with hardened weights, each could be shared within or between channels or layers.
        Alternatively, $K$-LUT inputs could be sacrificed to allow some or all weights to be stored in RAM and updated at runtime, enabling up-to cycle-by-cycle switches in inference operator behaviour.
        While both of these proposals would result in lower throughput and logic density---and necessitate more complex and time-consuming training---the implementation of larger LUTNet-based networks on smaller devices would become feasible.
        We will explore these in our future work.
        
        Fig.~\ref{plot:AREA_LUT_TRADEOFF} shows that while our expansion to 2-LUTs results in significant logic density gains over XNORs, returns for movement to $K$-LUTs for $K > 2$ are diminishing.
        We suspect that this is due to our current restriction on the form of function $g_n$ in \eqref{eq:synapse_lutnet}, \emph{i.e.} $\left\{-1,1\right\}^K \to \left\{-1,1\right\}$ rather than $\left\{-1,1\right\}^K \to \mathbb{N}$.
        This makes \eqref{eqn:retrain_solution} insoluble when $\hat{c}_{\boldsymbol{d}}$ is restricted to binary values.
        We can overcome this, and potentially make even more efficient use of the underlying FPGA fabric, by learning the popcount circuitry along with XNOR substitutes, replacing the summation as well as $w_n x_n$ in \eqref{eq:synapse_normal}.
        
        While the introduction of nonlinearity significantly increases the expressiveness of each inference operator, the experiments reported in Section~\ref{sec:eval_area_accuracy} revealed that 6-LUTNet showed signs of overfitting.
        In the future, we will explore methods of throttling expressiveness during training guided by losses, \emph{e.g.} switching to higher or lower $K$ when appropriate.
        
        Finally, LUTNet's software does not currently skip zeroes during training.
        As networks increase in size, GPU RAM will be increasingly inefficiently used, resulting in unnecessarily long training times.
        A future revision will therefore incorporate sparse matrix multiplication, preventing the storage of and multiplication by zero-valued weights.

    \section{Conclusion}
	
        In this paper, we introduced LUTNet: the first DNN architecture featuring $K$-LUTs as inference operators specifically designed to suit FPGA implementation.
        Our novel training approach results in the construction of $K$-LUT-based networks robust to high levels of pruning with little or no accuracy degradation, enabling the achievement of significantly higher area and energy efficiencies than that of traditional BNNs.
        
        In our experiments with 4-LUT-based inference operators, FPGA implementations following our proposals achieved a mean area reduction of 1.81$\times$ \emph{vs} the state-of-the-art BNN architecture with unrolling and pruning.
        These designs targetted a range of standard DNN models and datasets, required approximately the same training time and reached accuracies bounded within $\pm$0.300~pp in all cases.
        Due to their efficient use of soft logic, LUTNet implementations can exhibit energy efficiencies up to 6.66$\times$ greater than reported by the authors of related prior works.
        Thanks to its parameter hardening, our architecture also requires no use of block RAM: a common bottleneck for FPGA-deployed DNNs.
        
        The authors of existing works on low-precision DNN inference seem to have assumed that their forward-propagation functions must be good approximations of the linear dot product.
        With LUTNet, we argue for a tangential approach: through the embracement of nonlinearity, one can do more with less by unlocking the full potential of the $K$-LUT.
	
	\section*{Acknowledgements}
	
		The authors are grateful for the support of the United Kingdom EPSRC (grant number EP/P010040/1), Imagination Technologies and the Royal Academy of Engineering.
		
		Supporting data for this paper are available online at \texttt{https://doi.org/10.5281/zenodo.2616489}.
		
    \bibliographystyle{IEEEtran}
    \bibliography{pynq_cnn_bibliography} 

\end{document}